\pdfoutput=1

\documentclass[11pt]{article}

\usepackage[final]{acl}

\usepackage{times}
\usepackage{latexsym}

\usepackage[T1]{fontenc}

\usepackage[utf8]{inputenc}

\usepackage{microtype}

\usepackage{inconsolata}

\usepackage{booktabs}
\usepackage{bbm}
\usepackage{amsmath}
\usepackage{xcolor}
\usepackage{pifont}
\usepackage{graphicx}
\usepackage{soul} 
\usepackage{array}
\usepackage{tikz}
\usepackage{amssymb}
\usepackage{multirow}
\usepackage{float}
\usepackage{makecell}
\usepackage{enumitem}
\usepackage{hyperref}
\usepackage[normalem]{ulem}
\usepackage[capitalize]{cleveref}
\definecolor{LightGreen}{RGB}{230,250,245} 
\definecolor{DarkGreen}{RGB}{0,120,87}
\definecolor{LightOrange}{RGB}{255,240,235} 
\definecolor{DarkOrange}{RGB}{212,82,41}

\newcommand{\highlightgreen}[1]{\textcolor{DarkGreen}{\colorbox{LightGreen}{#1}}}
\newcommand{\highlightorange}[1]{\textcolor{DarkOrange}{\colorbox{LightOrange}{\sout{#1}}}}
\newcommand{\highlightorangew}[1]{\textcolor{DarkOrange}{\colorbox{LightOrange}{#1}}}
\newcommand{\redcross}{\textcolor{red}{\ding{55}}}
\newcommand{\greencheck}{\textcolor{green}{\ding{51}}}
\newcolumntype{P}[1]{>{\centering\arraybackslash}p{#1}}
\newcommand {\otoprule}{\midrule [\heavyrulewidth]}
\newcolumntype {+}{ >{\global\let\currentrowstyle\relax}}
\newcolumntype {^}{ >{\currentrowstyle }}
 \newcommand {\rowstyle}[1]{\gdef\currentrowstyle{#1} %
 #1\ignorespaces
 }
\newcommand{\tabhead}{\rowstyle{\bfseries}}
\newcommand{\Arrow}[1]{%
\parbox{#1}{\tikz{\draw[->](0,0)--(#1,0);}}
}
\title{CEval: A Benchmark for Evaluating Counterfactual Text Generation}

\author{Van Bach Nguyen \\
  University of Marburg, Germany \\
  \texttt{vanbach.nguyen@uni-marburg.de} \\\And
  Christin Seifert\\
  University of Marburg, Germany \\
   \texttt{christin.seifert@uni-marburg.de}
   \\\AND
   Jörg Schlötterer \\
  University of Marburg, Germany \\
  University of Mannheim, Germany \\
  \texttt{joerg.schloetterer@uni-marburg.de}
  }

\begin{document}
\maketitle
\begin{abstract}
Counterfactual text generation aims to minimally change a text, such that it is classified differently. Assessing progress in method development for counterfactual text generation is hindered by a non-uniform usage of data sets and metrics in related work. We propose CEval, a benchmark for comparing counterfactual text generation methods. CEval unifies counterfactual and text quality metrics, includes common counterfactual datasets with human annotations, standard baselines (MICE, GDBA, CREST) and the open-source language model LLAMA-2. Our experiments found no perfect method for generating counterfactual text. Methods that excel at counterfactual metrics often produce lower-quality text while LLMs with simple prompts generate high-quality text but struggle with counterfactual criteria. By making CEval available as an open-source Python library, we encourage the community to contribute additional methods and maintain consistent evaluation in future work.\footnote{\url{https://github.com/aix-group/CEval-Counterfactual-Generation-Benchmark}\label{gitlink}}

\end{abstract}

\section{Introduction}
\begin{figure}[t]
    \centering
    \includegraphics[width=\columnwidth]{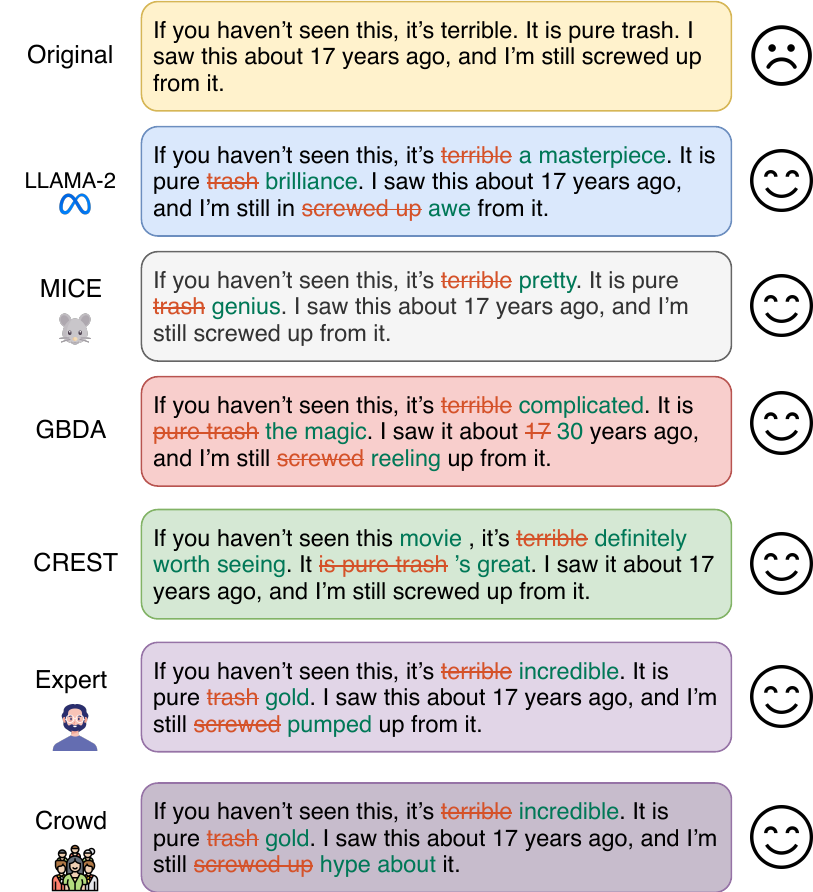}
    \caption{Examples of counterfactuals generated by different methods and human annotators that successfully flip the label from negative to positive for the same original instance.}
    \label{fig:intro}
\end{figure}

The rise of deep learning and complex ``black-box'' models has created a critical need for interpretability. As \citet{miller_explanation_2019} notes, explanations often involve counterfactuals to understand why event \textit{P} occurred instead of \textit{Q}. Ideally, these explanations show how minimal changes in an instance could lead to different outcomes. For example, to explain why the review \textit{``The film has funny moments and talented actors, \textbf{but it} feels long.''} is negative rather than positive, a counterfactual like \textit{``The film has funny moments and talented actors, \textbf{yet} feels \textbf{a bit} long.''} can be used (see \cref{fig:intro} for more counterfactual examples generated by different methods on the same original instance). This explanation highlights specific words to change and modifications needed for a positive sentiment . It also motivates counterfactual generation, which requires modifying an instance minimally to obtain a different model prediction.
\begin{table*}[t!]
\centering
\footnotesize
\begin{tabular}{p{3cm}p{3cm}p{3cm}p{5cm}}
\toprule \tabhead
\makecell{\textbf{Method}} & \textbf{Dataset} & \textbf{Metrics} & \textbf{Baseline} \\
\otoprule
\makecell{MICE\\\cite{ross_explaining_2021}} & \makecell[l]{IMDB, Race,\\Newgroups } & \makecell[l]{Flip rate, Fluency,\\ Minimality} & MICE's variants \\
\midrule
\makecell{CF-GAN\\\cite{robeer_generating_2021}}& \makecell[l]{HATESPEECH,\\ SST-2, SNLI} & \makecell[l]{Fidelity,\\Perceptibility,\\Naturalness}  &  \makecell[l]{SEDC~\cite{martens_explaining_2014}\\ PWWS+~\cite{ren_generating_2019}\\ Polyjuice~\cite{wu_polyjuice_2021}\\TextFooler~\cite{jin_is_2020}} \\
\midrule
\makecell{CORE \\\cite{dixit_core_2022}}& IMDB, MNLI & \makecell[l]{Diversity,\\ Closeness,\\ Accuracy} & \makecell[l]{Polyjuice~\cite{wu_polyjuice_2021}\\  GPT-3~\cite{brown2020language}\\ Human-CAD} \\
\midrule
\makecell{DISCO \\\cite{chen_disco_2023}}& SNLI, WANLI & \makecell[l]{Flip Score,\\ Diversity,\\ Accuracy} & \makecell[l]{Tailor~\cite{ross_tailor_2022}\\  Z-aug~\cite{wu-etal-2022-generating}\\ Human-CAD} \\
\bottomrule
\end{tabular}
\caption{Inconsistent use of datasets, metrics, and baselines across different methods. }
\label{tab:intro:moti}
\end{table*}
Besides explanations~\cite{robeer_generating_2021}, the NLP community uses counterfactuals for debugging models~\cite{ross_explaining_2021}, data augmentation~\cite{dixit_core_2022, chen_disco_2023, bhattacharjee2024llmguided}, and enhancing model robustness~\cite{treviso_crest_2023, wu_polyjuice_2021}. However, because it requires deciding where and how to change the text, with many possible modifications and a vast vocabulary. While many counterfactual generation methods for text data exist in the literature, they lack unified evaluation standards. \cref{tab:intro:moti} highlights inconsistencies in datasets, metrics, and baselines across different studies, making it difficult to compare different methods or selecting the most suitable method for specific applications. To overcome these limitations, a comprehensive benchmark to thoroughly evaluate counterfactual generation methods is necessary. 
A benchmark that provides standardized datasets, metrics, and baselines, enabling fair and effective comparisons, and ultimately driving progress in counterfactual generation.

This work introduces CEval, the first comprehensive benchmark for evaluating methods that modify text to change classifier predictions, including contrastive explanations, counterfactual generation, and adversarial attacks. CEval offers a robust set of metrics, incorporating established metrics from the literature alongside a novel metric we propose that captures probability changes rather than hard flip rates. This set enables the assessment of both ``counterfactual-ness'' (e.g., label flipping ability) and textual quality (e.g., fluency, grammar, coherence). The benchmark includes curated datasets with human annotations and a strong baseline using a large language model with a simple prompt to ensure high evaluation standards. Using CEval, we systematically review and compare state-of-the-art methods, highlighting their strengths and weaknesses in generating counterfactual text.
We analyze how automatically generated counterfactuals compare to human examples, revealing gaps and opportunities for improvement. We find that counterfactual generation methods often generate text that lacks in quality compared to simple prompt-based LLMs. 
In contrast, while the latter typically exhibit higher text quality, they may struggle to satisfy counterfactual metrics.
These insights suggest exploring combinations of both paradigms into hybrid methods as promising direction for future research. 
By demonstrating that an open-source LLM can serve as an alternative to a closed-source LLM in text evaluation, we make the benchmark completely open-source, thereby promoting reproducibility and facilitating further research in this domain.

\section{Related Work}\label{sec:related_work}
Terms like ``counterfactual'' and ``contrastive'' generation are often used interchangeably in literature~\cite{stepin_survey_2021} and our work adopts an inclusive definition. We define counterfactual generation as a process of generating a new instance $x'$, from the original instance $x$, that results in a different model prediction $y'$ with minimal changes. This definition includes counterfactual, contrastive generation, and adversarial attacks. Primarily, adversarial attacks focused on changing the label without considering text quality. Recent work like GBDA~\cite{guo_gradient-based_2021}  focuses on producing adversarial text that is more natural by adding fluency and semantic similarity losses. Hence, we include GBDA in our benchmark. Technically, counterfactual generation methods for text fall into three categories:\\
\textbf{Masking and Filling Methods (MF):}
These methods perform two steps: (1) identifying important words for masking by various techniques, such as selecting words with the highest gradient or training a separate rationalizer for the masking process and (2) replacing the masked words using a pretrained language model with fill-in-the-blank capability.
In step (1), MICE \cite{ross_explaining_2021} and AutoCAD~\cite{wen_autocad_2022} use the gradient of the classifier. DoCoGen~\cite{calderon_docogen_2022} identifies all domain-specific terms by calculating a masking score for n-grams (where n$\leq 3$) and masks all n-grams with a masking score exceeding a threshold $\tau$. Meanwhile, CREST~\cite{treviso_crest_2023} trains SPECTRA~\cite{guerreiro-martins-2021-spectra} as a separate rationalizer to detect which phrases or words to mask. In step (2), each of these methods fine-tunes T5 to fill in the blanks created during masking. Additionally, Polyjuice~\cite{wu_polyjuice_2021} takes text with user-specified manual masking as input and fine-tunes a RoBERTa-based model to fill in the blanks using control codes.\\
\textbf{Conditional Distribution Methods (CD):}
Methods like GBDA~\cite{guo_gradient-based_2021} and CF-GAN~\cite{robeer_generating_2021} learn a conditional distribution for counterfactuals. The counterfactuals are obtained by sampling from this distribution based on a target label.\\
\textbf{Counterfactual Generation with Large Language Models:}
Recently, there has been a trend towards using Large Language Models (LLMs) for counterfactual generation. Approaches like CORE~\cite{dixit_core_2022}, DISCO~\cite{chen_disco_2023} and FLARE~\cite{bhattacharjee2024llmguided} optimize prompts fed into LLMs to generate the desired counterfactuals. This trend is driven by the versatile capabilities of LLMs in various tasks~\cite{maynez-etal-2023-benchmarking}.


Despite the diverse approaches proposed in generating counterfactuals across various studies, the common objective remains to generate high-quality counterfactuals. However, previous studies employed different metrics, baselines, and datasets, as illustrated in \cref{tab:intro:moti}.  Therefore, given the rapid growth of approaches in this field, establishing a unified evaluation standard becomes paramount.
Existing benchmarks for counterfactual generation~\cite{pawelczyk_carla_2021, moreira_benchmarking_2022} focus exclusively on tabular data with properties that are orthogonal to text (e.g., continuous value ranges).
Hence, we introduce CEval to fill this gap and provide a standard evaluation framework specifically tailored to textual counterfactual generation. Our benchmark unifies metrics of both, counterfactual criteria and text quality assessment, including datasets with human annotations and a simple baseline from a large language model.




\section{Benchmark Design}
We focus on counterfactual generation for textual data, which involves editing given text with minimal modifications to produce new text that increases the probability of a predefined target label with respect to a black-box classifier. This process aims to generate a counterfactual, denoted as \( x' \), that changes the classifier's predictions compared to the original text \(x\).

Formally, given a fixed classifier \(f\) and a dataset with $N$ samples $(x_1, x_2, \ldots, x_N)$, $x_i = (z_1, z_2, \ldots, z_n)$ represents a sequence of \(n\) tokens. 
The original prediction is denoted as \(f(x) = y\), while the counterfactual prediction is \(y' \neq y\). 
The counterfactual generation process is represented by a method \(e : (z_1, \ldots, z_n) \mapsto (z'_1, \ldots, z'_m)\), ensuring that \(f(e(x)) = y'\). 
The resulting counterfactual example is \(x' = (z'_1, \ldots, z'_m)\) with \(m\) tokens.

A valid counterfactual instance should satisfy the following criteria~\cite{molnar2022}:\\
\textbf{Predictive Probability}: A counterfactual instance \(x'\) should closely produce the predefined prediction \(y'\). In other words, the counterfactual text should obtain the desired target label.\\
 \textbf{Textual Similarity}: A counterfactual \(x'\) should maintain as much similarity as possible to the original instance \(x\) in terms of text distance. This ensures that the generated text remains coherent and contextually aligned with the original.\\
  \textbf{Likelihood in Feature Space}: A counterfactual should exhibit feature values that resemble real-world text, indicating that \(x'\) remains close to a common distribution for text. 
  This criterion ensures that the generated text is plausible, realistic and consistent with typical language patterns. \\
\textbf{Diversity:}
When an explanation is ineffective, humans can offer alternatives. Similarly, if a counterfactual is unrealistic or not actionable, it is beneficial to modify the original instance differently to provide diverse options~\cite{mothilal_2020_dice}. Therefore, an effective counterfactual method should present multiple ways to change a text instance to obtain the target label. Diversity is measures for a set of counterfactual instances.

\subsection{Metrics}
In CEval, we use two types of metrics: \textit{counterfactual metrics}, which reflect the counterfactual criteria outlined above, and \textit{textual quality metrics}, which assess the quality of the generated text, irrespective of its counterfactual properties.
\subsubsection{Counterfactual metrics}
\textbf{Flip Rate (FR):} measures how effectively a method can change labels of instances with respect to a pretrained classifier. This metric represents the binary case of the \textit{Predictive Probability} criterion, determining whether the label changed or not and is commonly used in the literature~\cite{treviso_crest_2023,ross_explaining_2021}. FR is defined as the percentage of generated instances where the labels are flipped over the total number of instances $N$~\cite{bhattacharjee2024llmguided}:
\[FR =\frac{1}{N} \sum_{i=1}^{N} \mathbbm{1}[f(x_i) \neq f(x'_i)] \]
where $\mathbbm{1}$ is the indicator function.\\
\textbf{Probability Change ($\Delta \textbf{P}$):}
While the flip rate offers a binary assessment of \textit{Predictive Probability}, 
it does not capture the magnitude of change towards the desired prediction.
Some instances may get really close to the target prediction but still fail to flip the label. For example, a review such as: \textit{The movie looks great but has a confusing plot and slow pacing} is close to a positive label but remains negative. Consequently, its probability for the positive label should be larger than for a review like \textit{This movie is terrible}, which is really negative. The Probability Change ($\Delta$P) metric captures such cases by quantifying the difference between the probability of the target label $y'$ for the original instance \(x\) and the probability of the target label for the contrasting instance \(x'\).
\[ \Delta P =\frac{1}{N} \sum_{i=1}^{N}\bigl(P(y'_i \mid x'_i, f) - P( y'_i \mid x_i, f)\bigr)\]
Here, $P(y \mid x,f)$ is the probability that classifier $f$ assigns to label $y$ on instance $x$.\\
\textbf{Token Distance (TD):}
To measure \textit{Textual Similarity}, we use the token-level Levenshtein distance \(d(x, x')\) between the original instance \(x\) and the counterfactual \(x'\). This metric captures all types of text edits—insertions, deletions, and substitutions—making it ideal for evaluating minimal edits as counterfactual generation involves making these specific edits rather than completely rewriting the text. The Levenshtein distance is widely used in related work on counterfactual generation (e.g., \citet{ross_explaining_2021, treviso_crest_2023}).
\[TD= \frac{1}{N} \sum_{i=1}^{N} {d(x_i, x'_i})\]
\textbf{Perplexity (PPL):}
To evaluate whether the generated text is plausible, realistic, and follows a natural text distribution, we use perplexity from GPT-2 because of its effectiveness in capturing such distributions~\cite{radford_language_nodate}.\footnote{While we use GPT-2 in this study, any other LLM with strong text generation capabilities is a viable drop-in replacement.}
\[PPL(x) = \exp\left\{-\frac{1}{n} \sum_{i=1}^{n} \log p_{\theta}(z_i \mid z_{<i})\right\}\]
where \(\log p_{\theta}(z_i \mid z_{<i})\) is the log-likelihood of token \(z_i\) given the previous tokens \(z_{<i}\).\\
\textbf{Diversity (Div):}
We quantify diversity by measuring the token distance between pairs of generated counterfactuals. Given two counterfactuals, \(x'^1\) and \(x'^2\), for the same instance \(x\), diversity is defined as the average pairwise distance between the sets of counterfactuals:
\[ Div = \frac{1}{N} \sum_{i=1}^{N} d(x'^1_i, x'^2_i)\]
Here, \(d(x'^1_i, x'^2_i)\) is the Levenshtein distance between the corresponding tokens of the two counterfactuals for the \(i\)-th instance.

\subsubsection{Text Quality Metrics}
In addition to counterfactual evaluation metrics, we measure the quality of the generated text. \textit{Text quality metrics} are designed to evaluate specific aspects of texts. Following~\cite{chiang_can_2023, wang-etal-2023-chatgpt}, key text quality metrics for comprehensive insights into text quality are:
1) \textbf{Fluency} -- natural and readable text flow;
2) \textbf{Cohesiveness} -- logical and coherent structure and
3) \textbf{Grammar} -- syntactical and grammatical accuracy.

Combined with counterfactual metrics, text quality metrics provide a comprehensive view on effectiveness and linguistic quality of generated counterfactuals. 
Evaluating these text quality metrics usually requires human annotations, which are costly and time-consuming. Recently, \citet{chiang_can_2023, huang2023chatgpt, wang-etal-2023-chatgpt} showed that LLMs, specifically GPT-3/4 and ChatGPT, can serve as an alternative to human evaluation for assessing text quality using these metrics.
In this work, we use \textit{ChatGPT (gpt-3.5-turbo-0125)} with a temperature of 0.2 to evaluate the above textual quality metrics on a scale from 1 to 5 following~\cite{chiang_can_2023,gilardi2023chatgpt}.









\subsection{Datasets and Classifiers}
We chose two benchmark datasets for different NLP tasks: sentiment analysis on IMDB~\cite{maas_learning_2011} and natural language inference (NLI) on SNLI~\cite{bowman_large_2015}.
For both datasets, human-generated counterfactuals from crowdsourcing~\cite{kaushik_learning_2019} are available and for IMDB also from experts~\cite{gardner_evaluating_2020}. 
Additional datasets with pre-trained classifiers can be added to the benchmark.

IMDB contains diverse movie reviews from the IMDB website, along with corresponding sentiment labels (positive or negative) for each review. We selected the 488 instances with human-generated counterfactuals, balanced between 243 negative and 245 positive reviews~\cite{maynez-etal-2023-benchmarking}. Using a pre-trained BERT model\footnote{\url{https://huggingface.co/textattack/bert-base-uncased-imdb}} from TextAttack~\cite{morris-etal-2020-textattack} with 89\% accuracy, the counterfactual task is to minimally edit reviews to alter the classifier's prediction. 

SNLI~\cite{bowman_large_2015} consists of sentence pairs labeled as entailment, contradiction, or neutral, requiring models to understand semantic relationships. Using a pre-trained BERT model\footnote{\url{https://huggingface.co/textattack/bert-base-uncased-snli}} from TextAttack~\cite{morris-etal-2020-textattack} with 90\% accuracy, the counterfactual generation methods have to modify the premise or the hypothesis to change the classifier's label. 
\section{Counterfactual Methods Selection}
In this section, we briefly describe the counterfactual generation methods we evaluate with our benchmark. We selected at least one representative for each of the categories \emph{Masking and Filling (MF), Conditional Distribution (CD) and Large Language Models (LLMs)} (cf. \cref{sec:related_work}) based on the following criteria:
\begin{itemize}[noitemsep]
    \item The authors provide reproducible source code.
    \item The method is problem agnostic and can be applied to multiple text classification tasks.
    \item The method has access to the underlying text classifier.
\end{itemize}
We used the criteria \emph{reproducible code} and \emph{problem agnostic} as hard filters and \emph{access to the target classifier} as soft filter. A \emph{problem agnostic} method is versatile enough to generate counterfactuals for various types of classification problems (whereas methods like Polyjuice~\cite{wu_polyjuice_2021} or Tailor~\cite{ross_tailor_2022} require control codes, which limits their flexibility).
Methods without access to the target classifier are at disadvantage, as they have no information about the internals of the target classifier. Hence, wherever available, we opted for a method with access to the target classifier.
The selection based on these criteria (cf. details in Appendix, \cref{tab:method_compare}) resulted in MICE, GDBA, CREST and LLAMA-2 as representative counterfactual generation methods. We briefly describe them in the following.

\textbf{MICE}~\cite{ross_explaining_2021} is a contrastive explanation generation method. It trains an editor to fill masked tokens in a text so that the final text changes the original label. The tokens to be masked are chosen based on the highest gradients contributing to the predictions, and binary search is used to find the minimum number of tokens to mask. This method requires access to the classifier to verify the label internally, representing a counterfactual generation method.

\begin{table*}[h]
\centering
\addtolength{\tabcolsep}{-4pt} 
\footnotesize
\begin{tabular}{p{1cm}lcccccccccccc}
\toprule \tabhead
&&\multicolumn{6}{c}{\textbf{IMDB}}&\multicolumn{5}{c}{\textbf{SNLI}}\\
\cmidrule{3-8}\cmidrule{10-14}
\multicolumn{1}{c}{}  & & LLAMA-2      & MICE      & GBDA      & CREST      & Expert      & Crowd  &  & LLAMA-2     & MICE      & GBDA          & CREST & Crowd        \\
\otoprule
\multirow{5}{*}{\rotatebox{90}{\makecell{ \textbf{CF}\\ \textbf{Metrics}}}}&Flip Rate $\uparrow$   & 0.7 & \textbf{1.0} & 0.97 & 0.71 & 0.81 & 0.85 & &0.39 & 0.85& \textbf{0.94} & 0.39 &0.75 \\

&$\Delta$Probability $\uparrow$ & 0.69 & 0.91 & \textbf{0.96} & 0.70 & 0.80 & 0.84 & &0.33 & 0.65 & \textbf{0.86} & 0.10 & 0.64\\

&Perplexity $\downarrow$  & \textbf{41.3} & 62.1 & 84.1 & 44.7 & 56.2 & 52.4 & &\textbf{57.0} & 160 & 143 & 60.9 &72.1 \\

&Distance $\downarrow$   & 73.9 & 38.5 & 46.1 & 70.5 & 29.3 & \textbf{25.0} & & 6.15 & 5.64 & 4.85 & \textbf{3.53} & 4.06 \\

&Diversity $\uparrow$   & 61.6 & 48.4 & 47.6 & \textbf{86.6} & 38.7 & 38.7& &- & - & - & - &- \\

\midrule
\multirow{5}{*}{\rotatebox{90}{\makecell{\textbf{Text}\\\textbf{Quality}}}}&Grammar $\uparrow$ & 3.18 &  2.71 & 2.16 & 2.18 & 2.90 &2.92&& 3.68 & 3.33 & 2.29 & 2.71 & 3.58  \\
&Cohesiveness $\uparrow$ & 3.12  & 2.81 & 2.38 & 2.27 & 2.99 &2.95&& 3.61 & 3.31 & 2.03 & 2.74 & 3.60 \\
&        Fluency $\uparrow$ & 3.13 & 2.79 & 2.37 & 2.33 & 2.99 &2.92&& 3.59 & 3.33 & 2.17 & 2.70 & 3.56 \\

&\textit{Average} $\uparrow$    & \textbf{3.14}  & 2.77 & 2.30 & 2.27 & 2.96 &2.93&& \textbf{3.63} & 3.33 & 2.16 & 2.72 & 3.58 \\
\bottomrule
\end{tabular}
\addtolength{\tabcolsep}{4pt}
\caption{Results with counterfactual (CF) and text quality metrics on IMDB and SNLI. \textit{Average} denotes average of text quality metrics, each scored on a scale 1-5 following~\cite{chiang_can_2023}. We calculate diversity of the human groups by comparing expert with crowd counterfactuals and omit diversity on SNLI as it only has a single human counterfactual per instance (no expert annotations).}
\label{tab:result}
\end{table*}

\textbf{GBDA}~\cite{guo_gradient-based_2021} is a gradient-based adversarial attack that uses a novel adversarial distribution for end-to-end optimization of adversarial loss and fluency constraints via gradient descent. Similar to MICE, this approach needs access to the classifier for internal label verification. This method represents the adversarial attack domain.

\textbf{CREST}~\cite{treviso_crest_2023} follows a similar approach as MICE in first masking tokens that should be changed. 
Instead of using the highest gradient tokens to find the masks, the authors train a rationalizer using SPECTRA~\cite{guerreiro-martins-2021-spectra}.
Then, they fill the blanks with T5 same as MICE. Given the popularity of the Mask and Filling type, we chose this method for a more comprehensive comparison.

\textbf{LLAMA-2}~\cite{touvron2023llama}: Large Language Models have shown good performance on many tasks with only simple prompts~\cite{srivastava_beyond_2023}. Therefore, in this study, we use LLAMA-2 with simple one-shot learning as a baseline that is not specifically designed for counterfactual generation, but has strong language generation capabilities. The choice for LLAMA-2 as an open-source model is made in contrast to other studies that used closed-source LLMs.

The hyperparameters of each selected method can significantly impact the results, particularly for MICE~\cite{ross_explaining_2021} and CREST~\cite{treviso_crest_2023}. The percentage of masked tokens in both methods, representing the upper bound of changed tokens, directly influences the token distance and indirectly affects the flip rate: a lower percentage allows fewer tokens to change, resulting in a smaller distance but potentially a lower flip rate. In our experiments, we maintain the hyperparameters as specified in the original papers of each method. 
In case of LLAMA-2, the temperature of LLMs affects word sampling: lower temperatures yield more deterministic results, while higher temperatures enhance creativity. 
For the comparison with other methods, we use a temperature of 1.0 and analyze the impact of varying temperatures at the end of the next section.

\section{Results}

We evaluate all counterfactual generation methods against human crowd-sourced and human expert generations. Note that MICE and GBDA have access to the prediction model during generation, while CREST employs a pre-trained T5 model for internal label verification and transfers its prediction to the target BERT model. In contrast, LLAMA-2 and both human evaluation groups (crowd and expert) generate counterfactual examples solely based on the provided text and prompt.

We start with an example to illustrate the methods' varying characteristics before discussing our observations from the quantitative results.
\cref{fig:intro} shows the shortest example in the IMDB dataset where all methods, including human edits, change the label of the original sentence on the generated counterfactual.
For this simple instance, all methods and human groups agree on replacing negative words like \highlightorangew{terrible} and \highlightorangew{trash} with positive words, even though they differ in their choice of positive words. 
GDBA is the only exception, its replacements do not always convey a positive sentiment, which reduces text quality.
Similarly, MICE and CREST fail to detect the negative phrase \highlightorangew{screwed up}, which renders the text less cohesive and fluent than the text generated by LLAMA-2 and humans, who adapt this negative phrase as well.
Besides correctly identifying important words, GDBA also replaces irrelevant words like \highlightorangew{17}\Arrow{0.1cm}\highlightgreen{30}, resulting in a larger edit distance.
For a more complex example with higher variation of edits and generated text, see \cref{tab:negative_example} in the Appendix.


\paragraph{There is no single best method.}
\cref{tab:result} shows that no single method consistently outperforms the others, even on a single dataset. Methods with access to the target classifier, such as MICE and GDBA, excel at flipping the label but generate ``unnatural'' text with lower quality and higher perplexity due to poor grammar and low cohesiveness. In contrast, humans and LLAMA-2 consistently produce higher quality text across most metrics on both datasets. The lower success rate of humans in flipping the label suggests limitations in the target classifier, as perfect flip rates would be expected for human-generated text, the ``gold standard.'' Such potential issues are consistent with prior studies~\cite{kaushik_learning_2019,gardner_evaluating_2020}. 
Additionally, LLMs used as evaluation proxies, such as ChatGPT and GPT-2 (which measures perplexity), prefer LLAMA-2's output over human-generated text on both the SNLI and IMDB datasets. This preference is observed across different evaluator temperatures, as shown in \cref{tab:eval:compare_temp_model}, suggesting an interesting direction for further research into bias of LLMs as evaluators.

\paragraph{Diversity and distance are correlated.}
On the IMDB dataset, CREST and LLAMA-2 exhibit the highest diversity but also the highest distance. In contrast, human-generated changes (crowd and expert) are minimal and the least diverse. The Pearson correlation between diversity and distance is 0.93, indicating a very strong correlation between these two metrics. This strong correlation is likely due to minimal changes limiting the amount of variation. 

\paragraph{Probability changes are mostly bimodal.}
Interestingly, MICE has the highest flip rate (FR), but not the largest change in target label probability change ($\Delta$P) on the IMDB dataset. 
We observe a similar pattern when comparing LLAMA-2 and CREST on the SNLI dataset. CREST has an equal FR, despite LLAMA-2 inducing a larger $\Delta$P. 
\begin{figure}[t]
    \centering
    \includegraphics[width=\columnwidth]{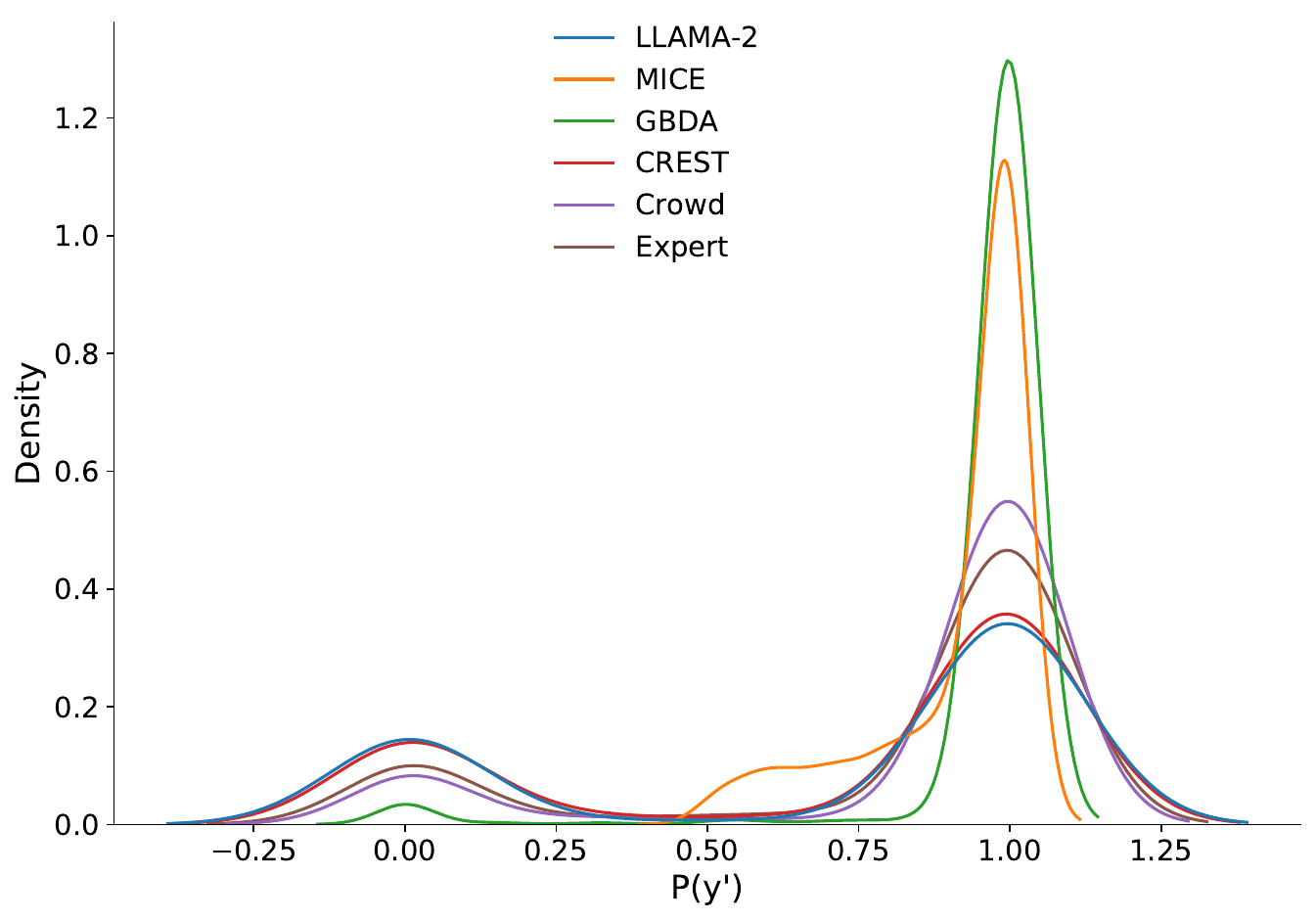}
    \caption{Distribution of target label probabilities of all methods on the IMDB dataset, including original text and human groups.}
    \label{fig:prob_y_distribution}
\end{figure}
A high FR combined with a low $\Delta$P suggests that the counterfactuals generated by the method are close to the decision boundary of the target classifier.   \cref{fig:prob_y_distribution} shows that only MICE generates a noticeable amount of instances that are close to the decision boundary ($P(y')=0.5$).
All others, including human groups, exhibit a bimodal pattern with narrow peaks at the two extremes.
While the imperfect FR of human groups suggests limitations in the target classifier, the distribution pattern may indicate the source of those limitations:
This pattern points to a poorly calibrated, overconfident target classifier, a common issue in today's deep learning architectures~\cite{guo2017calibration}.

\begin{figure}[htbp]
    \centering
    \includegraphics[width=\columnwidth]{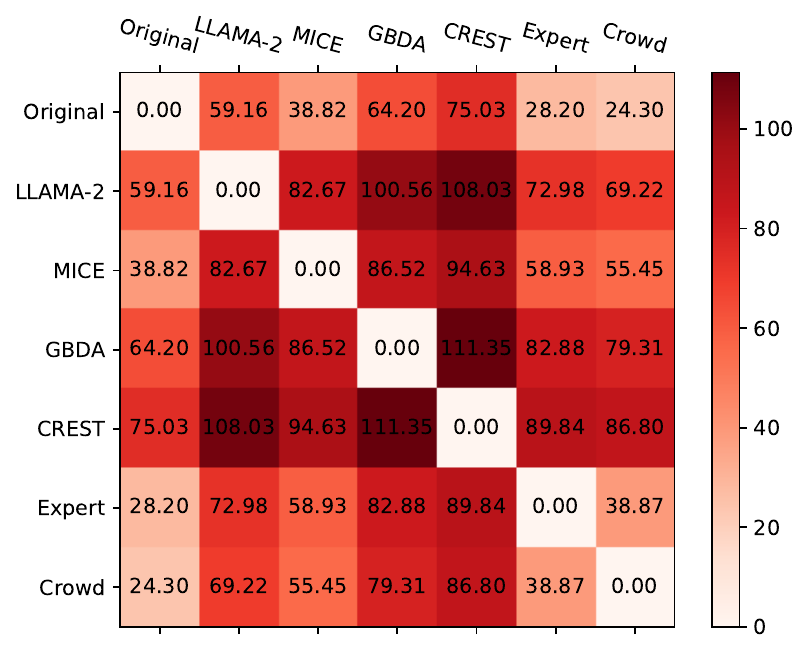}
    \caption{Avg. pairwise Levenshtein distance on IMDB.}
    \label{fig:pairwise_distances}
\end{figure}

\paragraph{Generated texts exhibit substantial differences.}
Among automatically generated methods, MICE's counterfactuals are closest to the original texts\footnote{In \cref{tab:result} we report distance only for true counterfactuals.} on the IMDB dataset, but still edit more tokens than humans (expert and crowd). The distance scores of CREST and LLAMA-2 are similar, as are those for MICE and GBDA, and for expert and crowd edits on the IMDB dataset. 
However, similar edit distances do not imply that these methods make the same edits.
To investigate the similarity of edits by different methods, we calculated the average pairwise distance between all generated examples on the IMDB dataset, regardless of label flip success.
The results are visualized in \cref{fig:pairwise_distances}. 
Crowd and expert edits are highly similar, indicating substantial overlap in their modifications. 
MICE generated text is closest to human edits, which makes it the most promising candidate to serve as proxy for human-generated counterfactuals.
GBDA and CREST have the largest distance to all other methods (including the original text) and to each other, i.e., their edits are largely distinct.
This substantial difference in generated texts suggests that robustness analyses of the target classifier should always be conducted with multiple methods.
\begin{table*}[t]
\centering
\footnotesize
\begin{tabular}{lcccccp{6pt}cccccp{6pt}cccccc}
\toprule \tabhead
&\multicolumn{5}{c}{\textbf{Grammar}}& &\multicolumn{5}{c}{\textbf{Cohesiveness}}& &\multicolumn{5}{c}{\textbf{Fluency}}\\
\cmidrule{2-6}\cmidrule{8-12}\cmidrule{14-18}
&\multicolumn{2}{c}{\textbf{GPT}} & &\multicolumn{2}{c}{\textbf{Mistral}}& &\multicolumn{2}{c}{\textbf{GPT}} & &\multicolumn{2}{c}{\textbf{Mistral}}& &\multicolumn{2}{c}{\textbf{GPT}} & &\multicolumn{2}{c}{\textbf{Mistral}}\\
&\textit{0.2} & \textit{1.0}  & & \textit{0.2} & \textit{1.0} &  & \textit{0.2} & \textit{1.0}  & & \textit{0.2} & \textit{1.0} &  & \textit{0.2} & \textit{1.0} &  & \textit{0.2} & \textit{1.0} \\
\cmidrule{2-3} \cmidrule{5-6} \cmidrule{8-9} \cmidrule{11-12} \cmidrule{14-15} \cmidrule{17-18}
Expert  & 2.90 & 2.94 &  & 4.81 & 4.74 &  & 2.99 & 2.99 &  & 4.74 & 4.66 &  &  2.99 & 2.99 &  & 3.91 & 3.91 \\
Crowd   & 2.92 & 2.89 &  & 4.88 & 4.79 &  & 2.95 & 2.98 &  & 4.78 & 4.68 &  &  2.92 & 2.94 &  & 3.83 & 3.81 \\
Crest   & 2.18 & 2.15 &  & 4.05 & 3.96 &  & 2.27 & 2.30 &  & 3.95 & 3.91 &  &  2.33 & 2.37 &  & 3.36 & 3.34 \\
GBDA    & 2.16 & 2.18 &  & 3.92 & 3.82 &  & 2.38 & 2.40 &  & 4.00 & 3.89 &  &  2.37 & 2.35 &  & 3.44 & 3.46 \\
Mice    & 2.71 & 2.73 &  & 4.55 & 4.44 &  & 2.81 & 2.82 &  & 4.40 & 4.35 &  &  2.79 & 2.81 &  & 3.77 & 3.75 \\
LLAMA-2 & \textbf{3.18} & \textbf{3.19}  & & \textbf{4.90} & \textbf{4.86}  & & \textbf{3.12} & \textbf{3.11} &  & \textbf{4.83} & \textbf{4.74}  & & \textbf{3.13} &\textbf{3.12}  & & \textbf{4.00} & \textbf{3.96 }\\
\bottomrule
\end{tabular}
\caption{Comparison of text quality evaluation using Mistral and ChatGPT (GPT-3.5 Turbo) with different temperatures (0.2 and 1.0) on IMDB dataset.}
\label{tab:eval:compare_temp_model}
\end{table*}
\paragraph{Temperature affects counterfactual generation diversity}
We compare LLAMA-2's temperature setting of 1.0 in \cref{tab:result} with additional values of 0.2 and 0.6 for \textit{counterfactual generation} and observe that the diversity score of LLAMA-2 varies significantly with temperature changes: the lower the temperature, the lower the diversity. For a temperature of 0.2, diversity score is 28.3 and for temperature 0.6, diversity score is 44.4 (details in Appendix, \cref{tab:gen:compare_temp_model}). This finding aligns with the expectation that higher temperatures, which increase token sampling flexibility, enhance the diversity of generated text. 
In contrast, other metrics remain largely unchanged or show minor variations.
For instance, average text quality is 3.15 at both temperatures of 0.6 and 0.2 on IMDB dataset.
\section{Comparison of LLMs for Text Quality Evaluation}
\label{sec:llm-comparison-evaluation}
Evaluating text quality with ChatGPT has been shown to be effective~\cite{huang2023chatgpt,gilardi2023chatgpt}. However, such evaluations come at high costs, limited control and customization constraints, and lack transparency. Therefore, we investigate an open-source LLM, Mistral-7B~\cite{jiang2023mistral} as an evaluation proxy.
\paragraph{Mistral-7B is a valid alternative to  ChatGPT}
To validate Mistral's evaluation capability, we use Mistral to evaluate the counterfactuals generated by all methods and compare the assessment scores with those from ChatGPT. Specifically, we compare the average scores, the Pearson correlation on the scores of each instance, and the Spearman correlation of the ranking of each method on all text quality metrics on both datasets and two temperature settings of 0.2 and 1.0. \cref{tab:eval:compare_temp_model} shows that Mistral-7B generally assigns higher scores than ChatGPT across all text quality metrics, though their scores are correlated. The Pearson correlation on the scores of each instance from the two models ranges from moderate to strong, with coefficients from 0.4 to 0.7, regardless of temperature settings (details in Appendix, \cref{fig:apdx:cor_eval_temp_model_imdb}). This implies that a text with high scores from Mistral is likely to receive high scores from ChatGPT as well. 
Furthermore, Spearman's rank correlation coefficients on the scores between the two models range from 0.89 to 1.0 , indicating a very strong correlation and partly even exactly identical rankings (details in Appendix \cref{tab:apdx:spear_corr}). 


To further validate Mistral-7B-instruct as a text quality evaluation proxy, we analyzed textual quality metrics on SNLI across two labels: contradiction and entailment. We hypothesized that entailment pairs exhibit higher cohesiveness and fluency than contradiction pairs, as entailment implies a logical relationship between the sentences. Our evaluation confirms that entailment pairs score significantly higher in text quality, particularly in cohesiveness and fluency, across all methods and human-generated texts. Detailed results are provided in Appendix, \cref{tab:llm_verification}.

Given the moderate to strong correlation with ChatGPT scores, very strong correlation in rankings and the validation of textual quality on the SNLI dataset, Mistral-7B is a viable alternative for comparative counterfactual method evaluation.
\paragraph{Text quality evaluation is robust to temperature variations}
Since temperature influences the performance of LLMs during inference~\cite{wang2023cost}, we evaluate its impact on their evaluation capabilities. Our study finds that text quality evaluation results are robust to temperature changes for both Mistral-7B and ChatGPT. We find a very strong correlation (Pearsons $\rho$ > 0.8) between evaluation scores for different temperatures of the same model (Appendix Figures~\ref{fig:apdx:cor_eval_temp_model_imdb} and \ref{fig:apdx:cor_eval_temp_model_snli}). Furthermore, the absolute scores remain similar across temperatures, as shown in \cref{tab:eval:compare_temp_model}.
\label{ssec:temp:gen}

\section{Conclusion}
We propose CEval to standardize the evaluation of counterfactual text generation, emphasizing the importance of both counterfactual metrics and text quality. Our benchmark facilitates standardized comparisons and analyzes the strengths and weaknesses of individual methods. Initial results show that counterfactual methods excel in counterfactual metrics but produce lower-quality text, while LLMs generate high-quality text but struggle to reliably flip labels. Combining these approaches could guide future research, such as using target classifier supervision to enhance LLM outputs. 
The diversity in method performance highlights the need for robustness analyses of target classifiers with multiple methods. 
Our findings also suggest that the target classifier may be poorly calibrated, warranting further investigation. 
Finally, we demonstrate that text quality evaluation using LLMs is robust to temperature changes. Additionally, we show that open-source LLMs, like Mistral, can serve as alternatives to closed-source models, such as ChatGPT, for evaluating text quality, thereby overcoming weaknesses of closed-source models, such as API deprecation or high costs.
This leads to CEval being a fully open-source Python library, encouraging the community to contribute additional methods and to ensure that future work follows the same standards. 
For future work, we plan to integrate LLMs specifically designed for evaluation, such as Prometheus~\cite{kim2023prometheus}, as an option for assessing text quality. Furthermore, instead of only considering the difference between instances to measure diversity, the diversity metric can be expanded to
incorporate the particular types of changes, such as negation and word replacements.

\section*{Limitations}
We employ default hyperparameters for each method and straightforward prompts with LLMs, which may not be optimal for the task at hand and could be further improved by hyperparameter optimization and prompt engineering.

This benchmark solely evaluates the quality of counterfactual text for explanation tasks. Further research is required to evaluate the performance of this text in other downstream tasks such as data augmentation with counterfactual examples or improving the robustness of the model using counterfactual examples. Additionally, we evaluate the metrics with a single BERT-based classifier. While this classifier achieves state-of-the-art classification accuracy, our results indicate that it might not be well calibrated. Estimating to which extent our findings can be generalized requires a combination of multiple diverse classifiers in the benchmark and the application in downstream tasks.

A potential exposure of ChatGPT or Mistral to the human counterfactual dataset is unlikely to impact our results, as we used these models only for evaluating text quality rather than counterfactual generation.
The exposure of LLAMA-2 to human counterfactuals remains uncertain. If such exposure occurred, it could potentially influence our results for LLAMA-2, as it would help to generate better (human-like) counterfactuals. However, \cref{fig:pairwise_distances} shows a considerable distance between human-generated and LLAMA-generated counterfactuals, suggesting a low likelihood of such influence.

\section*{Ethics Statement}
We use the publicly available datasets IMDB and SNLI, and employ the benchmark to evaluate existing counterfactual generation methods. None of these methods declared any ethical concerns. While the benchmark is designed to evaluate counterfactual generation methods to advance research in explainable AI, it could be  misused to select the best counterfactual methods for generating potentially harmful content.
One such harmful application scenario could be the generation of counterfactuals to evade a fake news detector. However, if such evasion would actually be possible without a drastic change of the semantics, the major risk stems from the counterfactual generation methods rather than from their benchmark comparison.

We strongly believe that a benchmark evaluation should be as open, fair, transparent and reproducible as possible. Therefore, we make all our source code (including benchmark evaluation and method implementation) publicly available\footref{gitlink} and include the option to evaluate text quality metrics with the open-source LLM Mistral-7B (cf. \cref{sec:llm-comparison-evaluation}).

\bibliography{custom}
\clearpage
\appendix

\section{Generated Text Comparison Example}
\cref{tab:negative_example} presents examples where the majority of methods were unsuccessful in altering the original label. While LLAMA-2 and human evaluators both identify \highlightorangew{nonsensical} words within the text, other methods overlook this aspect. In this intricate example, human crowdsource agreement with the human expert is not notably high, as their concurrence is limited to the term \highlightorangew{nonsensical}. However, the human expert's observations exhibit more alignment with other methods, such as modifying \highlightorangew{denigrate} akin to LLAMA-2, and replacing \highlightorangew{Sorry} or \highlightorangew{nonsense} as observed in MICE.
\section{Method Selection Criteria}
\label{sec:appendix}
\begin{table}[ht!]
\centering
\footnotesize
 \addtolength{\tabcolsep}{-3pt}
\begin{tabular}{p{2cm}cccc}

\toprule
 Method & Type & 
\makecell{Classifier\\ Access} & \makecell{Reproducible \\code}  & \makecell{Problem \\Agnosticity} \\
\otoprule
\textbf{MICE}   &MF  &\greencheck  &\greencheck      &\greencheck\\
CF-GAN &CD    &\greencheck  &\redcross        &\greencheck\\
Polyjuice          &MF  &\greencheck  &\greencheck    &\redcross   \\
\textbf{GBDA}               &CD    &\greencheck&\greencheck   &\greencheck \\
DISCO              &LLM   &\redcross&\redcross      &\greencheck \\
AutoCAD            &MF  &\greencheck&\redcross      &\greencheck \\
CORE               &MF  &\redcross&\redcross      &\redcross \\
DoCoGen            &MF  &\greencheck  &\greencheck  &\redcross  \\
Tailor~\cite{ross_tailor_2022}             &MF  &\greencheck  &\greencheck    &\redcross   \\
\textbf{CREST}              &MF  &\greencheck  &\greencheck      &\greencheck\\
GYC\cite{madaan_generate_2021}              &CD  &\greencheck  &\redcross     &\greencheck \\
FLARE               &LLM  &\redcross   &\redcross    &\greencheck  \\
\bottomrule
\end{tabular}
 \addtolength{\tabcolsep}{3pt}
\caption{Comparison of Methods. Methods of different types that meet  all inclusion criteria are highlighted in \textbf{bold} and are included in the benchmark.}
\label{tab:method_compare}
\end{table}

\section{Correlation of Mistral and ChatGPT }
\begin{table}[h!]
\centering

\begin{tabular}{lcc}
\toprule \tabhead
Temperature & \textbf{0.2} & \textbf{1.0} \\
\otoprule
Grammar & 1.0 & 0.89 \\
Cohesiveness & 0.94 & 0.89\\
Fluency & 1.0 & 0.94\\
\bottomrule
\end{tabular}
\caption{Spearman correlation of method rankings assigned by the LLM models Mistral and ChatGPT across different temperature settings, demonstrating very strong correlation.}
\label{tab:apdx:spear_corr}
\end{table}

\section{Effect of Temperature}
\begin{figure*}[t]
    \centering
    \includegraphics[width=\linewidth]{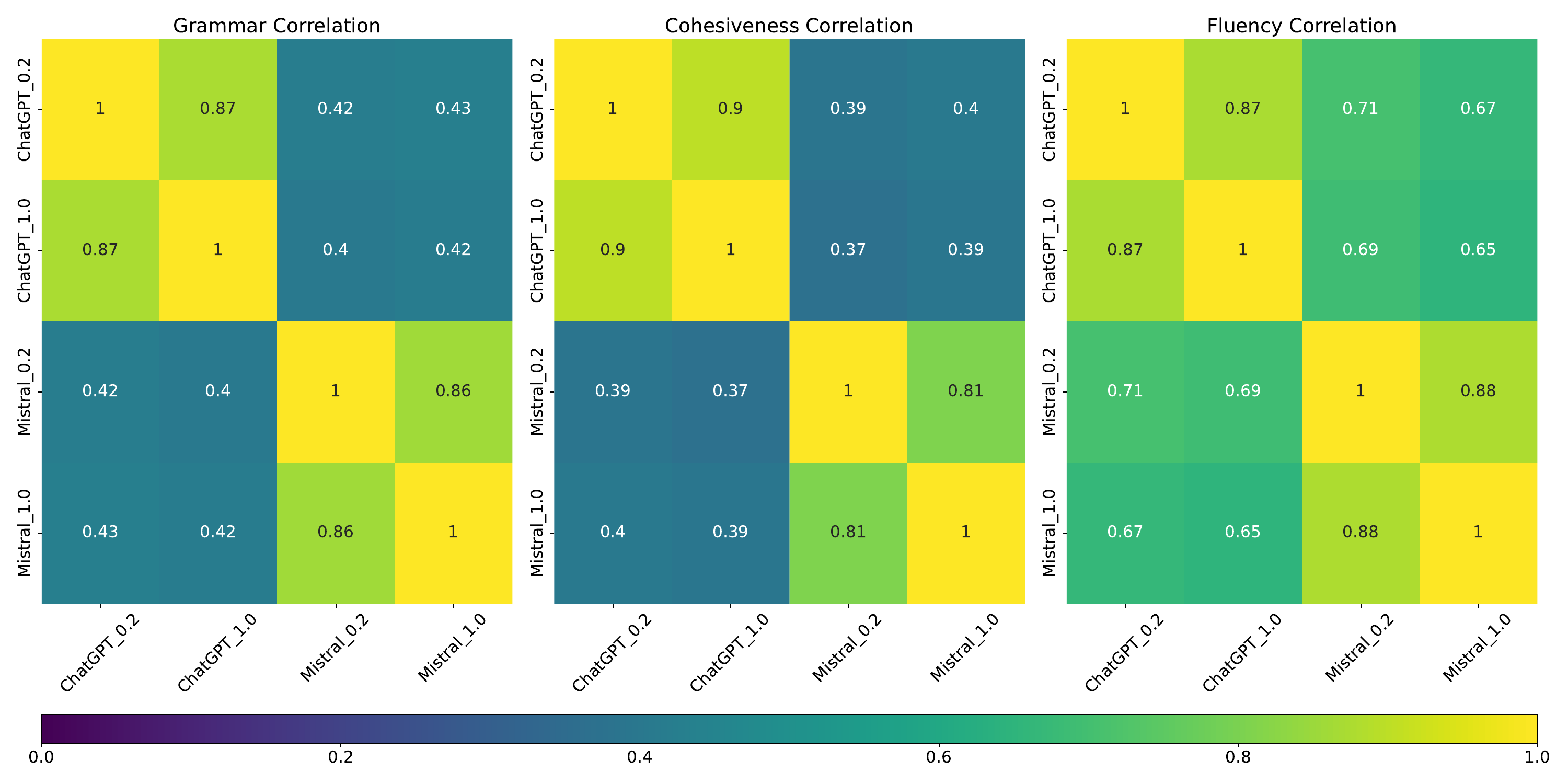}
    \caption{Pearson correlation between Mistral and ChatGPT in text quality evaluation with different temperatures (0.2 and 1.0) on the IMDB dataset. The same model with the different temperatures exhibits a strong correlation, meanwhile different models show a moderate correlation in evaluating text quality for counterfactual generation. }
    \label{fig:apdx:cor_eval_temp_model_imdb}
\end{figure*}

\begin{figure*}[t]
    \centering
    \includegraphics[width=\linewidth]{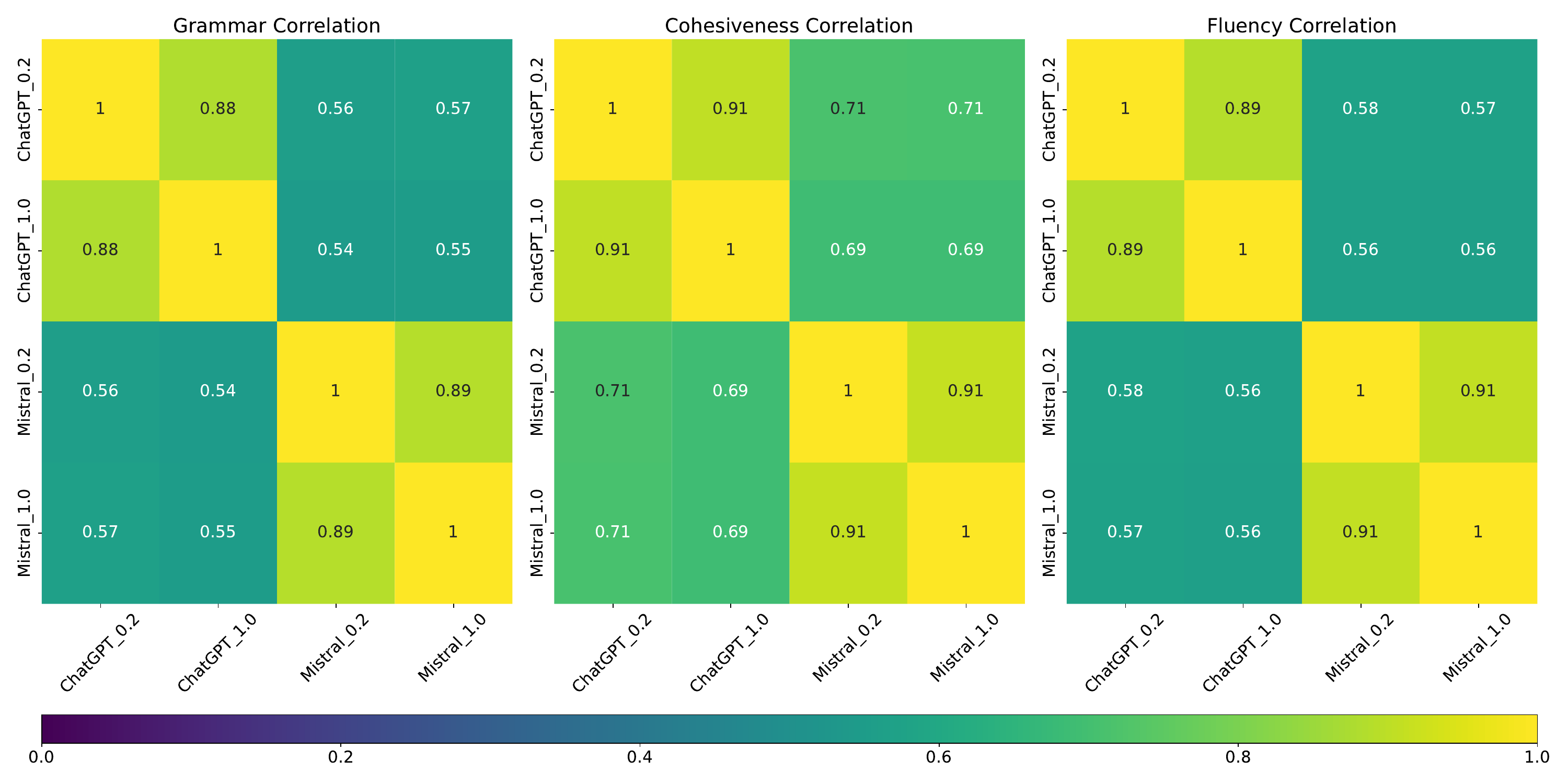}
    \caption{Pearson correlation between Mistral and ChatGPT in text quality evaluation with different temperatures (0.2 and 1.0) on the SNLI dataset. Text quality evaluation results of the same model with the different temperatures are strongly correlated; results from different models are moderately correlated.}
    \label{fig:apdx:cor_eval_temp_model_snli}
\end{figure*}

We evaluate the effect of temperature on the counterfactual generation process and text quality. \cref{tab:gen:compare_temp_model} shows the results of LLAMA-2 with three different temperatures: 0.2, 0.6, and 1.0. Lower temperatures imply a higher likelihood of selecting the most frequent tokens and a lower likelihood of selecting less frequent tokens. Consequently, diversity is low at lower temperatures and high at higher temperatures. Perplexity is also correlated with temperature, while other metrics do not show a clear correlation.
On the other hand, Figures~\ref{fig:apdx:cor_eval_temp_model_imdb} and \ref{fig:apdx:cor_eval_temp_model_snli} show the correlations between the same model at different temperatures, as well as the correlations between different models across various metrics. We observe a very strong correlation within the same model and a moderate correlation when using different models, suggesting that the evaluation is robust with respect to temperature.
\begin{table}[th!]
\centering
\addtolength{\tabcolsep}{-4pt}
\footnotesize
\begin{tabular}{p{1cm}lcccccccc}
\toprule \tabhead
&&\multicolumn{3}{c}{\textbf{IMDB}}&&\multicolumn{3}{c}{\textbf{SNLI}}\\
&& \textit{0.2}     & \textit{0.6}     & \textit{1.0}      && \textit{0.2}         & \textit{0.6}      & \textit{1.0}    \\
\cmidrule{3-5}\cmidrule{7-9}
\multirow{5}{*}{\rotatebox{90}{\makecell{ \textbf{CF}\\ \textbf{Metrics}}}}&Flip Rate $\uparrow$   & 0.68 & 0.65 & \textbf{0.70} && 0.38 & 0.40 & 0.39 \\

&$\Delta$Probability $\uparrow$ & 0.67 & 0.66 & \textbf{0.69} && 0.32 & \textbf{0.33} & \textbf{0.33}\\

&Perplexity $\downarrow$  & 40.6 & \textbf{39.1} & 41.3 && \textbf{54.9} &55.2 & 57.0 \\

&Distance $\downarrow$   & 50.7 & \textbf{48.9} & 58.0 && \textbf{4.36} & 4.48 & 4.78 \\

&Diversity $\uparrow$   & 28.3 & 44.4 & \textbf{61.6} && - & - & -\\

\midrule
\multirow{5}{*}{\rotatebox{90}{\makecell{\textbf{Text}\\\textbf{Quality}}}}&Grammar $\uparrow$ & \textbf{3.20} & 3.18 & 3.18 && 3.76 & \textbf{3.77} & 3.68 \\
&Cohesiveness $\uparrow$ & 3.14 & \textbf{3.15} & 3.12 && \textbf{3.71} & 3.69 & 3.61 \\
&        Fluency $\uparrow$ & 3.12 & 3.11 & \textbf{3.13} && 3.66 & \textbf{3.71} & 3.59\\

&\textit{Average} $\uparrow$    & \textbf{3.15} & 3.15 & 3.14 && 3.71& \textbf{3.72} & 3.63 \\
\bottomrule
\end{tabular}
\addtolength{\tabcolsep}{4pt}
\caption{Comparison of LLAMA-2 counterfactual generation with different temperatures (0.2, 0.6, and 1.0). Temperature primarily affects diversity, with minimal impact on other metrics.}
\label{tab:gen:compare_temp_model}
\end{table}

\begin{table*}[h!]
 \centering
 \footnotesize
 \addtolength{\tabcolsep}{-2pt}
\begin{tabular}{lccccccccccccccc}
\toprule \tabhead
& \multicolumn{3}{c}{\textbf{LLAMA-2}}      & \multicolumn{3}{c}{\textbf{MICE}}       & \multicolumn{3}{c}{\textbf{GBDA}}          & \multicolumn{3}{c}{\textbf{CREST}} & \multicolumn{3}{c}{\textbf{Crowd}}        \\
\otoprule
&\textit{E} & \textit{N}&\textit{C} &\textit{E} &\textit{N}& \textit{C} &\textit{E} &\textit{N}& \textit{C} & \textit{E} &\textit{N}& \textit{C} & \textit{E} &\textit{N}& \textit{C} \\
Grammar & 4.89 &\textbf{4.94}& 4.57 & \textbf{4.79} &4.67& 4.41 & \textbf{4.12} &4.00& 3.50 & \textbf{4.40} &3.84& 3.35 & \textbf{4.84} &4.84& 4.70\\
Cohesiveness & \textbf{4.29} &4.12& 2.01 & \textbf{4.26} &3.47& 2.31 & \textbf{2.86} &2.33& 1.58 & \textbf{3.19} &1.97& 1.55 & \textbf{4.08} &3.94& 3.06\\
Fluency & \textbf{4.99} & 4.86 & 4.38 & \textbf{4.90} &4.67& 4.38 & \textbf{4.61} &4.07& 3.56 & \textbf{4.43} &3.73& 3.13 & \textbf{4.95} &4.83& 4.30\\
\textit{Average} & \textbf{4.61} &4.50& 3.40 & \textbf{4.53} &4.06& 3.42 & \textbf{3.62} &3.20& 2.62 & \textbf{3.90} &2.96& 2.48 & \textbf{4.42} &4.33& 3.83\\
\bottomrule
\end{tabular}
\addtolength{\tabcolsep}{2pt}
\caption{Textual quality metrics to verify the LLMs evaluation. \textit{E}: Entailment, \textit{N}: Neutral, \textit{C}: Contradiction}
\label{tab:llm_verification}
\end{table*}

\begin{table*}[t]
\centering
\addtolength{\tabcolsep}{0pt} 
\footnotesize
\begin{tabular}{lcccccp{6pt}cccccp{6pt}cccccc}
\toprule \tabhead
&\multicolumn{5}{c}{\textbf{Grammar}}& &\multicolumn{5}{c}{\textbf{Cohesiveness}}& &\multicolumn{5}{c}{\textbf{Fluency}}\\
\cmidrule{2-6}\cmidrule{8-12}\cmidrule{14-18}
&\multicolumn{2}{c}{\textbf{GPT}} & &\multicolumn{2}{c}{\textbf{Mistral}}& &\multicolumn{2}{c}{\textbf{GPT}} & &\multicolumn{2}{c}{\textbf{Mistral}}& &\multicolumn{2}{c}{\textbf{GPT}} & &\multicolumn{2}{c}{\textbf{Mistral}}\\
&\textit{0.2} & \textit{1.0}  & & \textit{0.2} & \textit{1.0} &  & \textit{0.2} & \textit{1.0}  & & \textit{0.2} & \textit{1.0} &  & \textit{0.2} & \textit{1.0} &  & \textit{0.2} & \textit{1.0} \\
\cmidrule{2-3} \cmidrule{5-6} \cmidrule{8-9} \cmidrule{11-12} \cmidrule{14-15} \cmidrule{17-18}
Crowd & 3.58 & 3.56 && 4.62 & \textbf{4.61} && 3.60 & 3.53 && 3.77 & \textbf{3.73} && 3.56 & 3.51 && \textbf{4.48} & \textbf{4.43} \\
Crest & 2.71& 2.66 && 3.71 & 3.73 && 2.74 & 2.72 && 3.03 & 3.00 && 2.70 & 2.66 && 3.88 & 3.82 \\
GBDA & 2.29 & 2.31 && 3.27 & 3.22 && 2.03 & 2.08 && 2.10 & 2.20 &&  2.17 & 2.16 && 3.37 & 3.31 \\
Mice & 3.33 & 3.32 && 4.44 & 4.39 && 3.31 & 3.31 && 3.50 & 3.46 &&  3.33 & 3.34 && 4.38 & 4.29 \\
LLAMA-2 & \textbf{3.68} & \textbf{3.66} && \textbf{4.63} & 4.60 && \textbf{3.61} & \textbf{3.55} && 3.64 & 3.63 && \textbf{3.59} &\textbf{3.58} && 4.44 & 4.36 \\
\bottomrule
\end{tabular}
\addtolength{\tabcolsep}{0pt}
\caption{Comparison of text quality evaluation using Mistral and ChatGPT (GPT-3.5 Turbo) with different temperatures (0.2 and 1.0) on SNLI dataset.}
\label{tab:eval:compare_temp_model_snli}
\end{table*}


\begin{table*}[t]
\centering
\scriptsize
\begin{tabular}{lp{12cm}r}
\hline
\multirow{2}{*}{Method} & \multirow{2}{*}{Text} & Predicted \\
& & Label\\
\hline
Original & This movie frequently extrapolates quantum mechanics to justify nonsensical ideas, capped by such statements like "we all create our own reality".  Sorry, folks, reality is what true for all of us, not just the credulous.  The idea that "anything's possible" doesn't hold water on closer examination: if anything's possible, contrary things are thus possible and so nothing's possible. This leads to postmodernistic nonsense, which is nothing less than an attempt to denigrate established truths so that all ideas, well-founded and stupid, are equal.  To quote sci-fi writer Philip K. Dick, who put it so well, "Reality is that which, when you stop believing in it, doesn't go away.
 & Negative \\
\hline
LLAMA-2 & This movie frequently extrapolates quantum mechanics to justify \highlightorange{nonsensical}\highlightgreen{inspiring} ideas, capped by such statements like "we all create our own reality". Sorry, folks, reality is what true for all of us, not just the credulous. The idea that "anything's possible" doesn't hold water on closer examination: if anything's possible, contrary things are thus possible and so nothing's possible. This leads to postmodernistic nonsense, which is nothing less than an attempt to \highlightorange{denigrate}\highlightgreen{celebrate} established truths so that all ideas, well-founded and stupid, are equal. To quote sci-fi writer Philip K. Dick, who put it so well, "Reality is that which, when you stop believing in it, doesn't go away. & Negative\\
\hline
MICE & This \highlightorange{movie frequently extrapolates} \highlightgreen{excellent film has nothing more to say than to condemn} quantum mechanics to \highlightorange{justify}\highlightgreen{betray} nonsensical ideas, \highlightorange{capped}\highlightgreen{accompanied} by such statements like "we all create our own reality". \highlightorange{Sorry,}\highlightgreen{Hey,} folks, reality is what true for all of us, not just the credulous. The idea that "anything's possible" doesn't hold water on closer examination: if anything's possible, contrary things are thus possible and so nothing's possible. This \highlightorange{leads}\highlightgreen{movie is intended} to \highlightorange{postmodernistic nonsense, which} \highlightgreen{teach believers that embracing reality} is nothing \highlightorange{less} than an \highlightorange{attempt} \highlightgreen{excuse} to denigrate established truths so that all ideas, well-founded and \highlightorange{stupid}\highlightgreen{, doubtful}, are equal. To quote sci-fi writer Philip K. Dick, who put it so well, "Reality is that which, when you stop believing in it, doesn't go away. & Positive\\
\hline
GBDA & this movie \highlightorange{frequently} \highlightgreen{still} extrapolates \highlightorange{quantum} \highlightorange{mechanics} \highlightgreen{experimental} \highlightgreen{depression} to \highlightorange{justify} \highlightgreen{such} nonsensical ideas, \highlightorange{capped} \highlightgreen{accompanied} by \highlightorange{such} \highlightgreen{false} statements like like " we all create our own reality " . \highlightorange{sorry,} \highlightorange{folks, reality} \highlightgreen{".} \highlightgreen{nonetheless,}  \highlightgreen{nonetheless,} \highlightgreen{irony} is\highlightorange{what}\highlightorange{true}\highlightgreen{what,}  for all of us, not just the credulous. the idea that " anything's possible " doesn't \highlightorange{hold} \highlightorange{water} \highlightorange{on} \highlightorange{closer} \highlightorange{examination:} \highlightgreen{go} \highlightgreen{away} \highlightgreen{for} \highlightgreen{subjective} \highlightgreen{assumptions} \highlightgreen{:} if anything's possible, \highlightorange{contrary} \highlightgreen{everyday} things are \highlightorange{thus} \highlightgreen{ever} possible and so \highlightorange{nothing's} \highlightgreen{everything's} possible. this \highlightorange{leads} \highlightgreen{applies} to \highlightorange{postmodernistic}\highlightgreen{postmodernist} \highlightorange{nonsense,} \highlightgreen{authenticity,} which is nothing less than an attempt to denigrate \highlightorange{established} \highlightorange{truths} \highlightgreen{cultural} \highlightgreen{reality} so that \highlightorange{all} \highlightgreen{those} ideas, \highlightorange{well-founded} \highlightgreen{well} \highlightgreen{-} \highlightgreen{beautiful} and \highlightorange{stupid,} \highlightgreen{beautiful,} are \highlightorange{equal.} \highlightgreen{wonderful.} to quote \highlightorange{sci-fi} \highlightorange{writer} \highlightgreen{sci} \highlightgreen{-} \highlightgreen{fi} \highlightgreen{critic} philip k. dick, who \highlightorange{put} \highlightgreen{points} it so well, \highlightorange{"reality} \highlightgreen{"} \highlightgreen{comedy} is that which, when you stop believing in \highlightorange{it,} \highlightgreen{yourself,} doesn't go away. & Positive \\
\hline
CREST & \highlightorange{This} \highlightorange{movie} \highlightorange{frequently} \highlightorange{extrapolates} \highlightorange{quantum} \highlightorange{mechanics} \highlightorange{to} \highlightorange{justify} \highlightorange{nonsensical} \highlightorange{ideas,} \highlightorange{capped}\highlightgreen{A} \highlightgreen{quantum-sensical} \highlightgreen{thriller,} \highlightgreen{accompanied} by such \highlightorange{statements} \highlightgreen{films} like "we all create our own \highlightorange{reality"} \highlightgreen{world"}. Sorry, folks, \highlightgreen{this}\highlightorange{reality} is \highlightgreen{not} what true for all of us, not just the the \highlightorange{credulous}\highlightgreen{credulity}.The idea that "anything's possible" doesn't hold water on \highlightgreen{closer-to-end:} \highlightorange{closer} \highlightorange{examination:} if anything's possible, contrary things are thus possible and so \highlightorange{nothing's} \highlightgreen{that's} possible. \highlightorange{This} \highlightorange{leads} \highlightgreen{However,} \highlightgreen{there} \highlightgreen{is} \highlightgreen{no} \highlightgreen{less} \highlightgreen{reason} to \highlightgreen{definately} \highlightorange{postmodernistic} \highlightorange{nonsense,} \highlightorange{which} \highlightorange{is} \highlightorange{nothing} \highlightorange{less} \highlightorange{than} \highlightorange{an} \highlightorange{attempt} \highlightorange{to} \highlightorange{denigrate} established \highlightorange{truths} \highlightgreen{characters} so that \highlightorange{all} \highlightorange{ideas,} \highlightgreen{the} \highlightgreen{characters,} well-founded and \highlightorange{stupid,} \highlightgreen{well-meaning,} are \highlightorange{equal.} \highlightgreen{not.} To quote sci-fi writer Philip K. Dick, who put \highlightorange{it} \highlightorange{so} \highlightorange{well,} \highlightgreen{this} \highlightgreen{film} \highlightgreen{together,} \highlightorange{"Reality,} \highlightgreen{"Really,} \highlightorange{is} \highlightorange{that} \highlightorange{which,} when you stop believing in it, \highlightgreen{it} doesn't go away.& Negative \\
\hline
Expert & This movie frequently extrapolates quantum mechanics to justify \highlightorange{nonsensical} \highlightgreen{futurist} ideas, capped by such \highlightgreen{inspiring} statements like "we all create our own reality". \highlightorange{Sorry,} \highlightgreen{Yes,} folks, reality is \highlightgreen{this,} what true for all of us, \highlightgreen{is} \highlightgreen{what} \highlightgreen{we} \highlightgreen{just} \highlightgreen{see,} not just the credulous. The idea that "anything's possible" \highlightorange{doesn't} hold water \highlightgreen{even} on closer examination: if anything's possible, contrary things are thus possible and so nothing's \highlightorange{possible.} \highlightgreen{possible} \highlightgreen{but} \highlightgreen{we're} \highlightgreen{talking} \highlightgreen{alternate} \highlightgreen{universe.} This leads to postmodernistic \highlightorange{nonsense,} \highlightgreen{theories,} which \highlightorange{is} \highlightgreen{are} nothing less than an attempt to \highlightorange{denigrate} \highlightgreen{elevate} established truths so that all ideas, well-founded and stupid, are equal. To quote sci-fi writer Philip K. Dick, who put it so well, "Reality is that which, when you stop believing in it, doesn't go away. & Negative  \\
\hline
Crowd & This movie frequently extrapolates quantum mechanics to justify \highlightorange{nonsensical} \highlightgreen{wise} ideas, capped by such statements like "we all create our own reality". Sorry, folks, reality is what true for all of us, not just the credulous. The idea that "anything's possible" doesn't hold water on closer examination: if anything's possible, contrary things are thus possible and so nothing's possible. This leads to postmodernistic nonsense, which is nothing less than an attempt to denigrate established truths so that all ideas, well-founded and stupid, are equal. To quote sci-fi writer Philip K. Dick, who put it so well, "Reality is that which, when you stop believing in it, doesn't go away." \highlightgreen{This} \highlightgreen{movie} \highlightgreen{was} \highlightgreen{great} \highlightgreen{at} \highlightgreen{disputing} \highlightgreen{the} \highlightgreen{reality} \highlightgreen{of} \highlightgreen{things} \highlightgreen{and} \highlightgreen{I'd} \highlightgreen{recommend} \highlightgreen{it} \highlightgreen{for} \highlightgreen{everyone.} & Negative\\
\hline
\end{tabular}
\caption{Example for which most methods failed to flip the label}
\label{tab:negative_example}
\end{table*}
\end{document}